\documentclass{ieeeaccess}
\usepackage{cite}
\usepackage{amsmath,amssymb,amsfonts}
\usepackage{graphicx}
\usepackage{textcomp}
\usepackage{array}
\usepackage{booktabs}       
\usepackage{amsfonts}       
\usepackage{microtype}      
\usepackage{epsfig}
\usepackage{graphicx}
\usepackage{subfigure} 
\usepackage{floatrow}
\usepackage{multirow}
\usepackage{algorithm}
\usepackage{algpseudocode}
\usepackage{amsmath}

\def\BibTeX{{\rm B\kern-.05em{\sc i\kern-.025em b}\kern-.08em
    T\kern-.1667em\lower.7ex\hbox{E}\kern-.125emX}}
\begin{document}
\history{Date of publication xxxx 00, 0000, date of current version xxxx 00, 0000.}
\doi{10.1109/ACCESS.2017.DOI}

\title{RoIFusion: 3D Object Detection from LiDAR and Vision}
\author{\uppercase{Can Chen},
\uppercase{Luca Zanotti Fragonara, and Antonios Tsourdos}}
\address{Address: School of Aerospace, Transport and Manufacturing, Cranfield University, Cranfield, MK43 0AL, UK}


\markboth
{Author \headeretal: Preparation of Papers for IEEE TRANSACTIONS and JOURNALS}
{Author \headeretal: Preparation of Papers for IEEE TRANSACTIONS and JOURNALS}

\corresp{Corresponding author: Luca Zanotti Fragonara (e-mail: l.zanottifragonara@cranfield.ac.uk).}

\begin{abstract}
When localizing and detecting 3D objects for autonomous driving scenes, obtaining information from multiple sensor (e.g. camera, LIDAR) typically increases the robustness of 3D detectors. However, the efficient and effective fusion of different features captured from LIDAR and camera is still challenging, especially due to the sparsity and irregularity of point cloud distributions. This notwithstanding, point clouds offer useful complementary information. In this paper, we would like to leverage the advantages of LIDAR and camera sensors by proposing a deep neural network architecture for the fusion and the efficient detection of 3D objects by identifying their corresponding 3D bounding boxes with orientation. In order to achieve this task, instead of densely combining the point-wise feature of the point cloud and the related pixel features, we propose a novel fusion algorithm by projecting a set of 3D Region of Interests (RoIs) from the point clouds to the 2D RoIs of the corresponding the images. Finally, we demonstrate that our deep fusion approach achieves state-of-the-art performance on the KITTI 3D object detection challenging benchmark.
\end{abstract}


\titlepgskip=-15pt

\maketitle

\section{Introduction}

Object detection of 3D bounding boxes is one of the fundamental challenges of situational awareness and 3D environmental perception for autonomous systems (e.g. autonomous vehicles, robots, unmanned aerial vehicles, etc.). In fact, autonomous systems need to perceive objects in their surrounding environment using different sensors (e.g. cameras, LIDAR) for navigation and obstacle avoidance. In the past few years, 2D object detection for computer vision \cite{ren2015faster, redmon2016you, liu2016ssd, dai2016r, lin2017feature, lin2017focal, duan2019centernet} has made significant progresses, especially with the advent of Convolutional Neural Network (CNN) technology \cite{krizhevsky2014imagenet}. However, 3D object detection remains an open challenge, especially when multiple sensors are used to obtain a more reliable and robust information.

Recently, many researchers focused on the exploitation of LIDAR-only methods for 3D object detection due to the advantages that the point clouds provide precise depth information and dense geometric shape feature \cite{qi2017pointnet, qi2017pointnet++, wang2019dynamic, liu2019relation, chen2019gapnet, chen2020go}. \textit{PointRCNN} \cite{shi2019pointrcnn} builds a two-stage architecture to directly process dense 3D point clouds and estimate 3D bounding boxes from all the foreground points. \textit{VoxelNet} \cite{zhou2018voxelnet}, \textit{SECOND} \cite{yan2018second} convert the point clouds to voxels before applying standard CNNs to achieve the same result. \textit{Pixor} \cite{yang2018pixor}, \textit{Complex-YOLO} \cite{simon2018complex}, and \textit{Birdnet} \cite{beltran2018birdnet} operate deep CNNs on Bird-Eye-View (BEV) maps for 3D object classification and bounding box regression.

However, a standard point cloud is incapable of offering texture information and high resolution of an object, which are actually beneficial to capture discriminative features. In contrast, the images provide rich color and texture information, but with a lack of depth and scale information without the application of complex and computationally intensive algorithms (i.e. stereography). For example, small objects (e.g. pedestrians) detected at long-distance generates only few points in the point cloud, which makes the classification or localization of these objects very difficult with only a LIDAR. Meanwhile, in the image domain, texture and color features of small objects can be still visible, due to the higher spatial resolution of images, and likely to be captured by existing mature 2D CNNs technology. As a result, the fused features, leveraging the advantages from both point clouds and images, are beneficial in exploiting more reliable representations and improving the performance of the 3D object detection architecture.

However, it is still challenging to develop an efficient and effective sensor fusion method due to the viewpoint misalignment caused by the properties of the point clouds and the images. In order to address this issue, early method \textit{MV3D} \cite{chen2017multi} and \textit{AVOD} \cite{ku2018joint} perform 3D bounding box regression on fused 2D images and 2D Bird-Eye-View (BEV) feature maps, although quantization for BEV generation gives rise to a lot of geometric information losses. \textit{Frustum Pointnets}~\cite{qi2018frustum} and \textit{Pointfusion}~\cite{xu2018pointfusion} project 2D bounding boxes from the image-based 2D detector onto the point clouds to coarsely cluster potential foreground points. At this point, the PointNets are applied for 3D boxes estimation, but this procedure heavily relies on the performance of the 2D detectors. \textit{PointPainting} \cite{vora2019pointpainting} feeds the pixel-wise semantic features captured from image-based semantic segmentation model onto corresponding point-wise semantic features in the point cloud to boost the performance of the 3D object detection.

It can be observed that the main disadvantage of dense point-pixel fusion method \cite{vora2019pointpainting} is that they are leading to a considerable amount of redundant computations. Meanwhile, using a BEV-image fusion method allows the deep learning-based fusion of the feature maps captured from an individual viewpoint but with geometric information losses. However, it is the authors assumption that it is not strictly necessary to densely fuse the whole point clouds with images. Conversely, it is feasible to generate a small set of potential Region of Interests (RoIs), followed by the application of a deep fusion method only on those local regions used for 3D object detection. The advantages of this fusion method are that it considerably reduces the computation cost and allows an easy alignment of the viewpoints on the local regions.

Motivated by these observations, we hereby present an efficient and lightweight deep fusion method for 3D object detection for point clouds and images. Our main contributions can be summarized as follows:

\begin{itemize}
\item We propose a lightweight deep fusion neural network, named \textit{RoIFusion}, aiming at efficiently and effectively fusing the point clouds and the images for 3D object detection.
\item We propose a keypoints generation layer for the estimation of the keypoints on the objects guided by the fusion of the point clouds and the images, followed by a voting layer used to generate the center points of the objects.
\item We propose a RoIFusion layer to aggregate the 3D RoIs generated from the center points with the corresponding 2D RoIs which are obtained by projecting the 3D RoIs to the images. 
\item We evaluated our model on the KITTI dataset~\cite{geiger2013vision} and achieved state-of-the-art results compared with respect to other outstanding methods.
\end{itemize}

\section{Related Work}

\subsection{LIDAR-only methods for 3D object detection} 
Many existing methods explored the possibility of detecting the objects with 3D bounding boxes only using point clouds, as they provide accurate geometric information. It is possible to broadly classify these methods into four subcategories: projection-based methods, volumetric-based methods, pointnets-based methods, and point-voxel methods.

\paragraph{Projection-based methods} Several works \cite{simony2018complex, beltran2018birdnet, yang2018pixor} apply 2D CNNs directly to Bird-Eye-View (BEV) projected from the raw point clouds in order to estimate the 3D bounding box and orientation of an object. \textit{FVNet} \cite{zhou2019fvnet} projects the raw point clouds to the front view, which is then fed to a proposal generation network and a refinement network to estimate the parameters of the 3D bounding box (i.e. object location, size, and orientation). This method allows for building a lightweight neural network for real-time applications. However, it ignores the size and the location of the objects and suffers from lots of geometric information losses during quantization. As a result, it is unlikely to exploit sufficient discriminative features for 3D object detection.

\paragraph{Volumetric-based methods} Volumetric-based methods convert the raw point clouds to standard 3D grids and represents the point clouds as voxels. For instance, \textit{VoxelNet} \cite{zhou2018voxelnet} learns discriminative voxel-wise features for 3D region proposal generation and then proceeds to solve the 3D bounding box regression problem. However, many empty voxels are generated during the voxelization process, which leads to large computational cost because of the processing of those empty cells. In order to address this problem, \textit{SECOND} \cite{yan2018second} improved \textit{VoxelNet} \cite{zhou2018voxelnet} by proposing an efficient method, named \textit{sparse convolution} \cite{graham2017submanifold}, to ignore the empty voxels. Finally, \textit{PointPillars} \cite{lang2019pointpillars} converts the raw point clouds to a set of stacked pillars and then encodes the same to 2D pseudo-images, which can be used as input for 2D CNNs for 3D bounding box regression.

\paragraph{Pointnets-based methods} \textit{PointNets} \cite{qi2017pointnet, qi2017pointnet++} models are efficient in the exploitation of point cloud features. \textit{PointRCNN} \cite{shi2019pointrcnn} sets an example in the classification and regression of 3D bounding box directly from dense point clouds. Specifically, it firstly applies \textit{PointNet++} \cite{qi2017pointnet++} to extract dense semantic features for all the points, and then generates 3D region proposals for all the foreground points. Successively, the second stage is applied to refine predictions. However, the dense processing leads to quite heavy computational costs.

\paragraph{Point-voxel methods} In order to achieve high detection performance but to also reduce the computational costs, several works \cite{shi2020points, chen2019fast, liu2019tanet, shi2019pv} introduced two-stages neural networks for 3D object detection. In the first stage, they coarsely localise the objects and estimate the parameters of the bounding box from the voxel grids generated from the raw point clouds. In the second stage, they introduce a refinement module that leverages the \textit{PointNets} to refine the 3D bounding box. The methods leverage the sparsity of the voxel grids and the ability of \textit{PointNets} to carry out feature extraction and to gradually detect the 3D objects starting from coarse to more refined representations.

Finally, a recent one-stage method, \textit{3DSSD} \cite{yang20203dssd} abandons the refinement stage and builds a one-stage anchor-free neural network to directly regress 3D bounding box from the estimated candidate 3D RoIs.

\subsection{Image-only methods for 3D object detection}

In the past few years, 2D object detection has made great progress. However, estimating 3D bounding box directly from 2D images is still quite difficult due to the lack of depth information in single camera images. \textit{Mousavian et al.} \cite{mousavian20173d} estimates the pose and 3D bounding box by learning the geometric constraints from the 2D bounding box. \textit{Wang et al.} applies LIDAR-only 3D detectors on the Pseudo-LiDAR representations converted from the estimated image-based depth maps. \textit{Stereo R-CNN} \cite{li2019stereo} applies \textit{Faster R-CNN} \cite{ren2015faster} on both the left and right images and predicts 3D bounding boxes by learning the projection relations between the associated 2D left-right bounding boxes and 3D bounding box corners.

\subsection{Multi-sensor fusion methods for 3D object detection}

In order to get the best of both worlds, there are several works attempting to fuse point clouds and 2D images with various strategies. Early works such as \textit{MV3D} \cite{chen2017multi} and \textit{AVOD} \cite{ku2018joint} firstly used off-the-shelf 2D feature extractors to capture the feature maps from the images and the multi-view representations of the point clouds (e.g. Bird Eye View and Front View), which are then typically fused together by a sum or a concatenation operation. A Region Proposal Network (RPN) is then applied to the fused feature maps to generate 3D bounding box proposals, followed by a refinement network for final 3D bounding box prediction. The advantages of this method are that mature 2D object detector and 2D feature extractor technologies are available to be applied on the multi-view representations of the point clouds. Furthermore, the features from different sensors can interact over the stacked layers, as these features are normally obtained from similar or even the same neural networks. \textit{Liang et al.} \cite{liang2018deep} utilizes the continuous convolution method to fuse the feature maps of the images and BEVs. Specifically, this approach proposes a continuous fusion layer that aggregates each pixel feature in the image feature maps with the features of the neighbouring points in the BEV feature maps to learn a fused local region, which allows to extract sufficient discriminative features for 3D object detection.

In order to narrow the searching space, \textit{Frustum pointnets} \cite{qi2018frustum} and \textit{Frustum convnet} \cite{wang2019frustum} introduced the concept of 3D bounding frustums. The 2D bounding boxes are obtained from mature 2D detectors, and then the 3D frustums are used to trim the point cloud data. Finally, \textit{Pointnets} methods are applied to the trimmed point clouds for carrying out the 3D bounding box regression task. Similarly, \textit{Pointfusion} \cite{xu2018pointfusion} aggregates the global features of the image obtained from an off-the-shelf 2D feature extractor with the dense semantic features of the point cloud, which are captured from  \textit{Pointnet} \cite{qi2017pointnet}.

Finally, \textit{PointPainting} \cite{vora2019pointpainting} densely aggregates the output of the image segmentation neural network with the point clouds before applying LIDAR-only 3D detectors to boost the performance of the 3D object detection task.

\begin{figure}[!t]
  \centering
   {\epsfig{file = 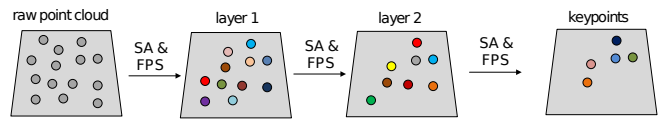, width=1.0\linewidth}}
  \caption{Illustration for point-guided keypoints generation}
  \label{fig:pointgk}
 \end{figure}
\unskip

\begin{figure*}[!t]
  \centering
   {\epsfig{file = 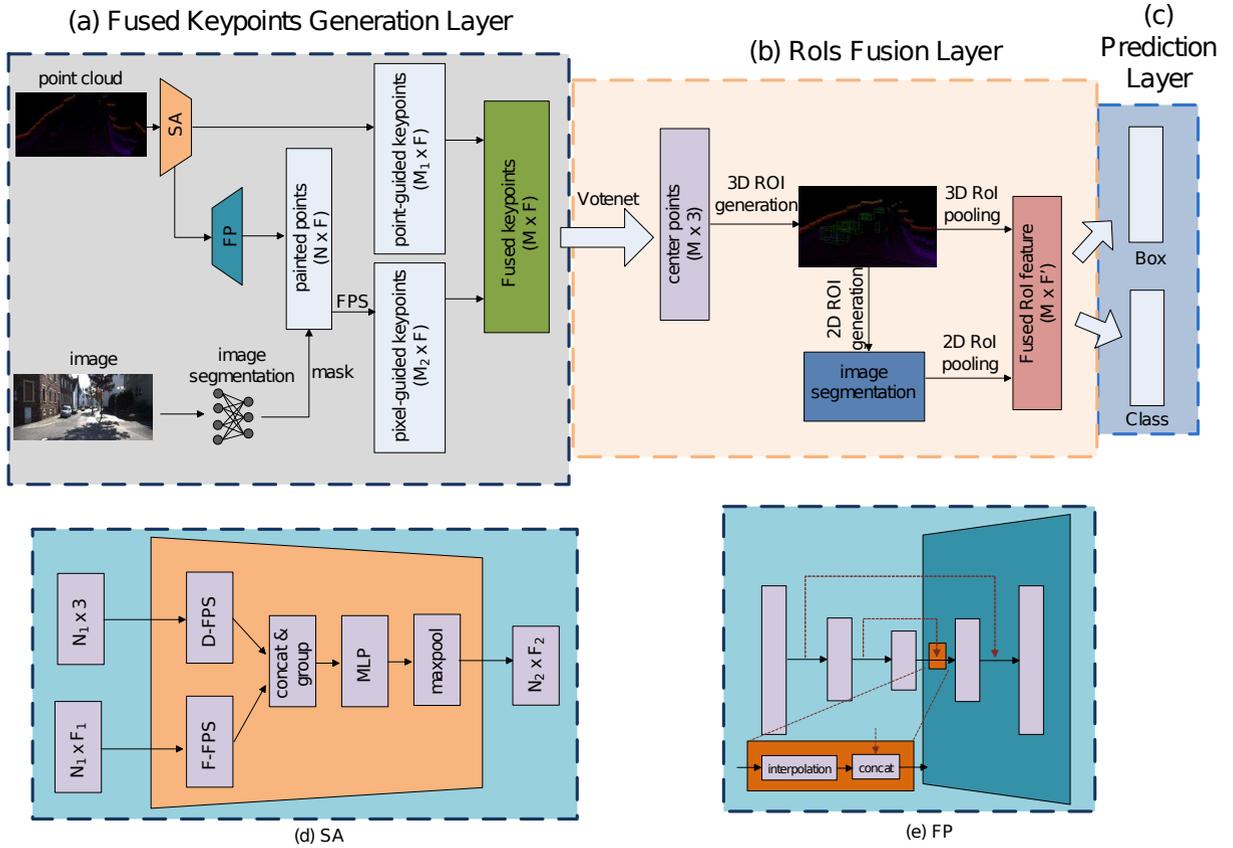, width=0.9\linewidth}}
  \caption{\textbf{RoIFusion framework.} As a whole, the proposed architecture contains three parts: (a) Fused Keypoints Generation (FKG) layer takes the raw point clouds and the images as input and aggregates the keypoints generated from both the point cloud set abstract (SA) network and the image segmentation network respectively. (b) The keypoints are then used to estimate the center points of the potential objects and the 3D Region of Interests (RoIs), which are projected to the image to obtain the 2D RoIs. The 3D/2D RoI pooling layers are employed to capture the respective local features, which are finally fused together for 3D bounding box prediction as shown in part (c). As shown in (d), the input of the SA module are the 3D points \(N_1 \times 3\) and the corresponding features \(N_1 \times F_1\). We simultaneously downsample the points and extract the corresponding deep features in the orange block, and obtain the downsampled 3D points \(N_1 \times 3\) and corresponding features \(N_2 \times F_2\). \(N_1,F_1,N_2,F_2\) are the dimension number of the input points/features, output points/features respectively. Part (e) illustrates the Feature Propagation (FP) module: the upsampling method that takes the output of the SA module as input and generates the segmented features for all the points in the point cloud.}
  \label{fig:structure}
 \end{figure*}
\unskip

\section{RoIFusion architecture}

In this section, we introduce and describe our RoIFusion neural network for 3D object detection as shown in Fig.~\ref{fig:structure}, which uses both raw point clouds and 2D images as input. Our goal is to leverage the fusion information captured from both sensor modalities to classify and localize the objects within the oriented 3D bounding boxes. In particular, we firstly propose a fused keypoints generation layer (FKG layer) to estimate a set of 3D keypoints from the point clouds, followed by a RoIs fusion layer to fuse the 3D RoI features in the point clouds with the 2D RoI features in the images by the 3D/2D RoI pooling operation respectively. At last, a prediction layer is proposed to predict the parameters of the oriented 3D bounding box.

\subsection{Fused Keypoints Generation (FKG) Layer}
Instead of densely generating 3D region proposals relying on all the foreground points, we only estimate a small set of 3D keypoints on the objects to generate the RoIs for deep fusion. As illustrated in part (a) of Fig.~\ref{fig:structure}, our FKG layer takes the raw point clouds and the RGB images as input, and combines point-guided keypoints and pixel-guided keypoints that are generated by  individually performing the LIDAR-only point cloud segmentation model and image-only segmentation network respectively. As a result, we obtain a set of keypoints on the objects leveraging both the point cloud and the image information.

We can define the point cloud as shown in Eq.~\eqref{eq:pc_eq}, where \(\mathbf{x_i}\) denotes the \(i\)-th point with the 3D space coordinates \([x_i,y_i,z_i]\) and the measured reflectance \(r_i\). As a result, the dimension of the point cloud data set is \(N \times 4\).
\begin{equation}\label{eq:pc_eq}
\mathbf{X}=\left\{ \mathbf{x}_i=[x_i,y_i,z_i,r_i] \in \mathbb{R}^4, i=1,2,\ldots,N\right\}
\end{equation}

\paragraph{Point-guided keypoints generation} Our point-guided keypoints \(\mathbf{\hat{X}^{(pc)}} \in \mathbb{R}^{M_1,F}\), where \(M_1, F\) are the number of keypoints and corresponding features respectively, are extracted by the set abstraction (SA) layers backbone, as illustrated in part (d) of Fig.~\ref{fig:structure}. This is done as proposed by \textit{PointNet++} \cite{qi2017pointnet++}, where a simultaneously downsampling of the points and the extraction of the corresponding deep features are carried out.

Specifically, as shown in Fig.~\ref{fig:pointgk}, we apply a number of SA layers with downsampling operation to the raw point cloud. At each layer, a set of points are processed and a new set with higher-level but fewer points is generated. Finally, we obtain a small number of points that are treated as keypoints.

With respect to the downsampling strategy, we use an iterative farthest point sampling (FPS) method to select the points for the subset. Let us suppose an empty subset \(\mathbf{X1}\), a random point is firstly picked and added to \(\mathbf{X1}\), then the point having the farthest 3D geometric Euclidean distance is iteratively added to \(\mathbf{X1}\) until the expected \(M\) points are picked. The FPS strategy, named \textit{D-FPS}, has a better coverage of the whole point set than random sampling. In order to preserve sufficient foreground points and filter out the background, inspired by \textit{3DSSD} \cite{yang20203dssd}, we decided to also employ a specific FPS strategy, named \textit{F-FPS}, which calculates the Euclidean distances of the semantic features for the points selection. The \textit{F-FPS} method is beneficial to preserving foreground points (e.g. points on the objects) and removing the useless background, such as points on the ground. Finally, we follow \cite{yang20203dssd} and combined both FPS strategies together to efficiently capture sufficient foreground points as the keypoints.

\begin{figure}[!t]
  \centering
   {\epsfig{file = 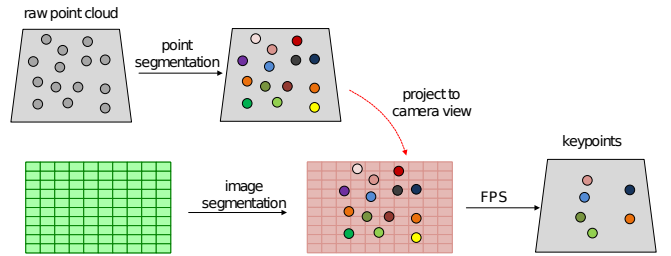, width=1.0\linewidth}}
  \caption{Illustration for pixel-guided keypoints generation}
  \label{fig:pixelgk}
 \end{figure}
\unskip

\paragraph{Pixel-guided keypoints generation} Considering the fact that the colour and texture representations are useful to localize objects within point clouds, especially for small objects that are difficult to be detected by LIDAR-only detectors, we capture the segmentation features and corresponding scores using an image segmentation network, which then is used to guide the keypoints selection as shown in Fig.~\ref{fig:pixelgk}.

The detailed procedure to extract the pixel-guided keypoints is shown in Alg.~\ref{alg:alg1}.Firstly, we generate the point clouds segmentation features \(\mathbf{X_s}\) using a Feature Propagation (FP) layer as shown in part (e) of Fig.~\ref{fig:structure}. In particular, we leverage the output of the SA layer as input, and upsample the points by interpolating the point features using the inverse squared Euclidean distance weighted average function as shown in Eq.~\eqref{eq:interpolation}. Furthermore, we concatenate the interpolated point features with the skip linked point features from the corresponding SA layer. As a result, our FP layer outputs the 3D geometric points and corresponding semantic features with the same number of points as the raw point cloud.

\begin{equation}\label{eq:interpolation}
f(\mathbf{x})=\frac{\sum_{i=1}^{k}\omega_i(\mathbf{x})f_i }{\sum_{i=1}^{k}\omega_i(\mathbf{x})}
\end{equation}
where \(\omega_i(\mathbf{x})=\frac{1}{(\mathbf{x-x_i})^2}\) is the inverse squared Euclidean distance between a certain point \(\mathbf{x}\) and corresponding \(i\)-th neighbouring point \(\mathbf{x_i}\) of the \(k=3\) nearest neighbours.

For what concerns the image processing, it is a common choice in literature to use mature 2D feature extractors to capture the feature maps from the RGB images. However, these feature maps are unlikely to localize the objects in the images. As a result, we use a lightweight image segmentation neural network \textit{DeepLabv3} \cite{chen2017rethinking} to efficiently capture pixel-wise segmentation features \(\mathbf{I_s}\) and segmentation scores \(\mathbf{S}\), which allows to ignore the background and conduct the keypoints selection. It is worth pointing out that our RoIFusion model is agnostic to the development of the image segmentation models.

After that, we project the point cloud to the image viewpoint to paint the point cloud with the segmentation features and the segmentation scores from the relevant pixels, then we mask the painted point cloud \(\mathbf{X_{img}}\) with the foreground image segmentation scores and map to the point clouds segmentation features to generate all the foreground segmentation features \(\mathbf{X_s^{(obj)}}\) for the point cloud. At last, we use the F-FPS as our downsampling strategy to further select a small set of point segmentation features \(\mathbf{\hat{X}^{(img)}}\) with the dimension size of \(M_2 \times F\), where \(M_2\) and \(F\) are the number of keypoints and corresponding features respectively.

\begin{algorithm}[htb]
  \caption{ Pixel-guided keypoints generation.}
  \label{alg:alg1}
  \begin{algorithmic}[1]
    \Require
      The point clouds \(\mathbf{X} \in \mathbb{R}^{N,4}\).

      The images \(\mathbf{I} \in \mathbb{R}^{W,H,3}\).

      Homogenous transformation matrix \(\mathbf{T} \in \mathbb{R}^{4,4}\).

      Camera projection matrix \(\mathbf{M} \in \mathbb{R}^{3,4}\).

    \Ensure 
      Pixel-guided keypoints features \(\mathbf{\hat{X}^{(img)}} \in \mathbb{R}^{M_2,F}\).
    \State  Apply for point cloud segmentation network to obtain segmentation features \(\mathbf{X_s} \in \mathbb{R}^{F_p}\).
   \State  Apply for image segmentation network to obtain segmentation features \(\mathbf{I_s} \in \mathbb{R}^{W,H,F_i}\) and segmentation scores \(\mathbf{S} \in \mathbb{R}^{W,H,C}\).
    \State  \(\mathbf{X_{img}}=Projection(\mathbf{M,T,X})\).
    \State  \(\mathbf{X_{obj}, index}=Mask(\mathbf{X_{img}, S})\).
    \State  \(\mathbf{X_s^{(obj)}}=Mapping(\mathbf{X_s, index})\).
    \State  \(\mathbf{\hat{X}^{(img)}}=FPS(\mathbf{X_s^{(obj)}})\). \\
    \Return \(\mathbf{\hat{X}^{(img)}}\).
  \end{algorithmic}
\end{algorithm}

\begin{figure}[h]
  \centering
   {\epsfig{file = 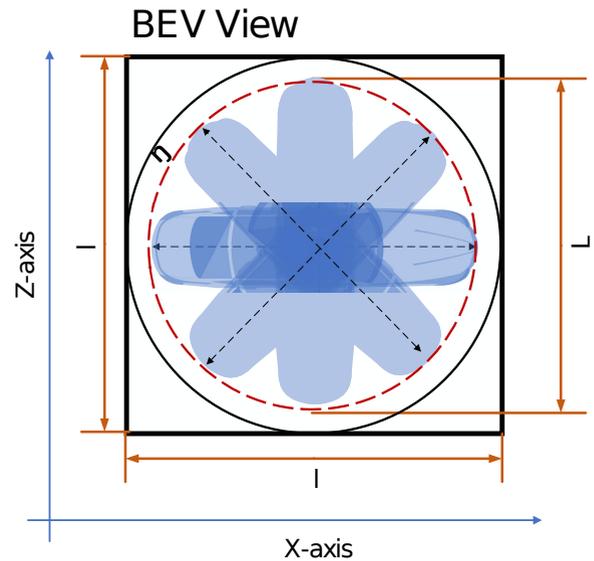, width=0.9\linewidth}}
  \caption{RoIs generation layer. The objects with all the oriented angles are illustrated in the red dot circle. The length size of the objects (e.g. cars) are defined as \(L\). \(\eta\) is the enlarged size for the object length. \(l\) represents the size of the RoI in the Bird eye view (BEV).}
  \label{fig:roigen}
 \end{figure}
\unskip

\paragraph{Keypoints fusion} We finally obtain the fused keypoints \(\mathbf{\hat{X}} \in \mathbb{R}^{M,F}\) by aggregating the point-guided keypoints with the pixel-guided keypoints along with the channel of the number of the keypoints, where \(M\) is the number of fused keypoints. We note that the points on small objects are likely to be selected due to the fact that a part of the points are captured based on the image segmentation scores.

\begin{figure*}[!t]
  \centering
   \subfigure[SA layer]{\label{fig:sa}\includegraphics[width=0.45\linewidth]{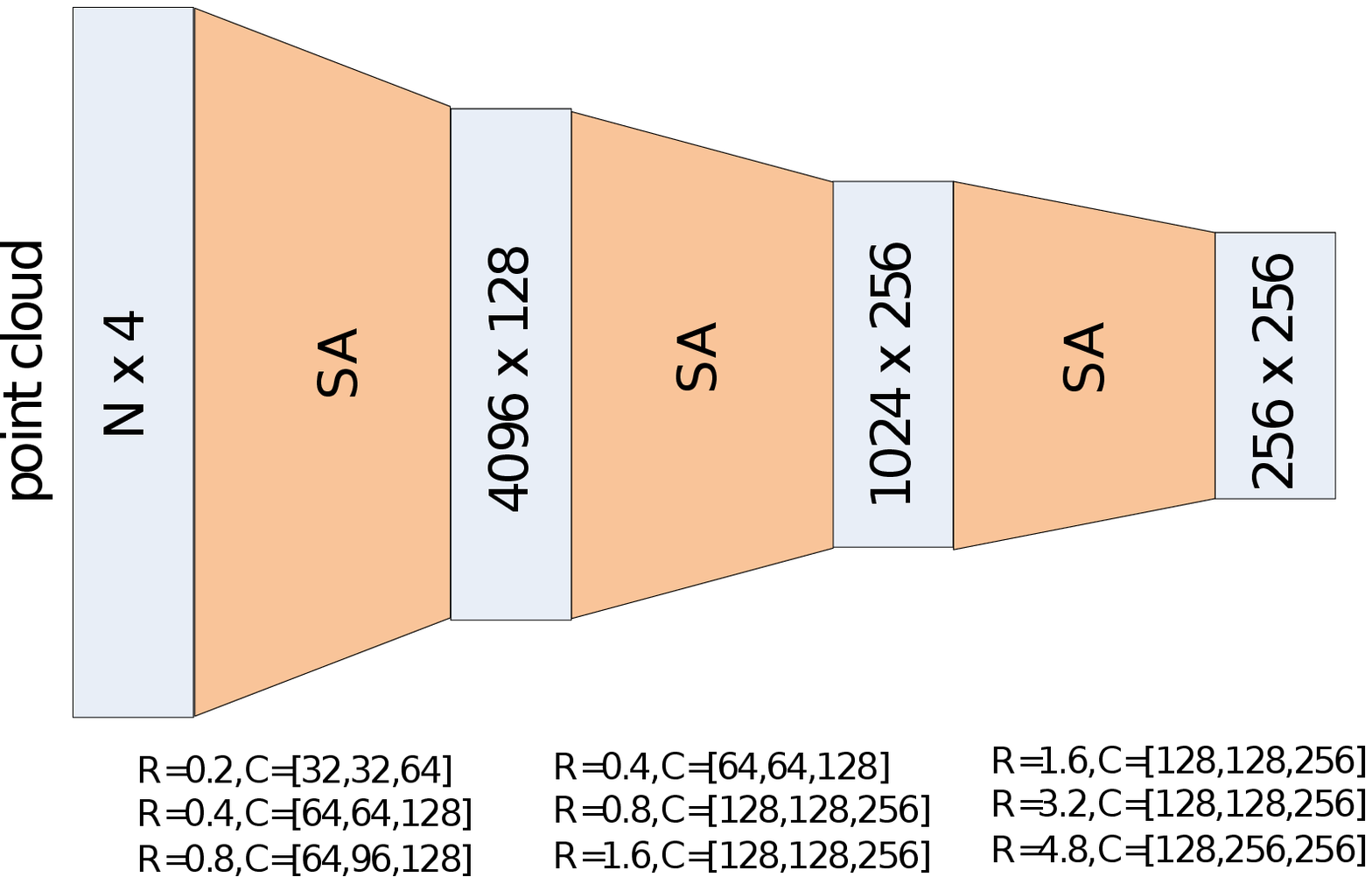}}
   \subfigure[FP layer]{\label{fig:fp}\includegraphics[width=0.5\linewidth]{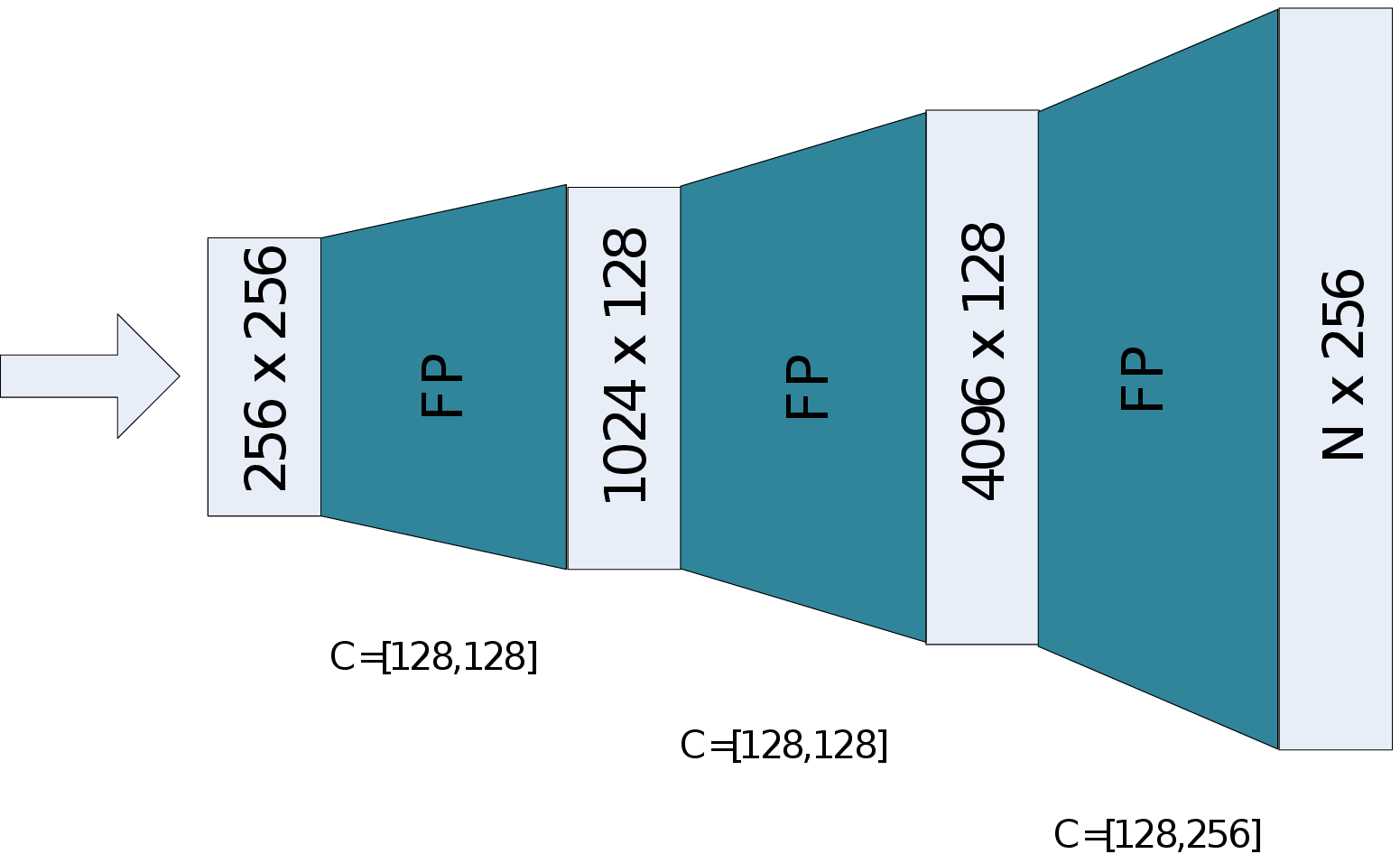}}
  \caption{Illustration of the SA and FP layers. The SA layer \ref{fig:sa} takes \(N\) points as input, followed by stacked SA layers to downsample points and extract corresponding features. After that, the FP layer \ref{fig:fp} upsamples the points to the original 16384 points using stacked FP layers. \(R\) is the radius of the ball query strategy for clustering local region points, and \(C\) is the number of the filters for the MLP layers.}
  \label{fig:safp}
 \end{figure*}
\unskip

\subsection{RoIs fusion layer}

After the implementation of the FKG layer intertwined with the point cloud segmentation network and the image segmentation network, we obtain a set of keypoints scattered over the objects, which are also used to predict the center of the objects before we generate the RoIs. Considering that these keypoints are on the objects, inspired by \cite{ding2019votenet}, we use the spatial location and features of the keypoints to estimate the corresponding center of the objects. As shown in part (b) of Fig.~\ref{fig:structure}, these high-level keypoints are used to generate 3D RoIs in the point cloud view and corresponding 2D RoIs in the camera view, followed by a RoI fusion operation to obtain the fused RoI features for further bounding box regression and object classification. Specifically, we build a subnetwork \textit{VoteNet} with a single layer to learn the spatial offset between predicted center points and corresponding ground truth. We treat each center point as the centroid of the 3D bounding box of the object.

\begin{figure}[h]
  \centering
   {\epsfig{file = 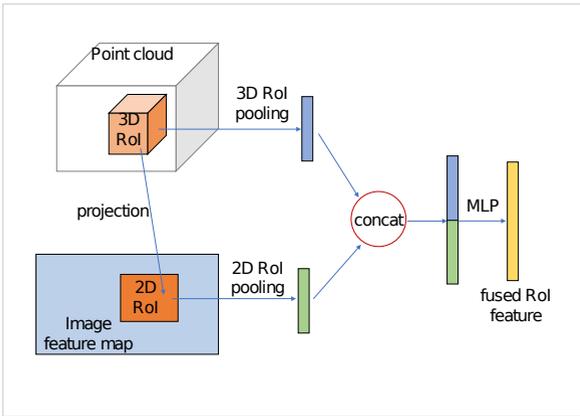, width=0.9\linewidth}}
  \caption{RoIs fusion layer}
  \label{fig:roifusion}
 \end{figure}
\unskip

\paragraph{3D RoIs generation and pooling} The 3D RoIs are generated for the center points we previously obtained. Successively, we apply a 3D RoI pooling layer to pool the surrounding points of each center point and learn the local features for those clustered points around each center point.

We encode our RoIs using the axis-aligned 3D bounding boxes. Specifically, the centroid of each RoI \((x^{(c)},y^{(c)},z^{(c)}\) is parametrized using the obtained center point. The length \(l\) and the width \(w\) of the RoI are set to the enlarged length size of the objects to cover all the orientation scenarios as shown in Fig.~\ref{fig:roigen}. We finally use an enlarged height of the objects as the height \(h\) for each RoI. As a result, the dimension of the RoI is defined as \((x^{(c)}_i,y^{(c)}_i,z^{(c)}_i, \eta+h_i, \eta+w_i, \eta+l_i)\), where \(\eta\) is the parameter for extended size of the RoI.

\begin{table}[!]
  \caption{KITTI dataset difficulty classification levels for object detection.}
  \label{tab:levels} \centering
  \begin{tabular}{p{1.2cm}p{1.5cm}p{2.1cm}p{1.2cm}}
    \toprule[0.96pt]
     Levels  &  \begin{tabular}[c]{@{}c@{}}Min. \\ bounding \\  box height \end{tabular}  & \begin{tabular}[c]{@{}c@{}}Max. \\ occlusion \\  level \end{tabular}  & \begin{tabular}[c]{@{}c@{}}Max. \\ truncation \end{tabular}  \\
    \midrule
    Easy              & 40 pixels                & Fully visible            & 15\%               \\
    Moderate              & 25 pixels                & Partly occluded            & 30\%               \\
    Hard              & 25 pixels                & Difficult to see            & 50\%               \\
    \bottomrule[0.96pt]
  \end{tabular}
\end{table}

After that, we shift the points inside each 3D RoI to the relative locations based on the center points for better local features learning and then apply a subnetwork equipped with stacked Multi-Layer-Perceptron (MLP) layers on the cluster the points inside the 3D RoIs to extract the local RoI pooling features.

\paragraph{2D RoIs generation and pooling} Our 3D RoIs are then projected to the image to generate the corresponding 2D RoIs, followed by a 2D RoI pooling layer, inspired by \cite{ren2015faster}, to learn the local texture features for the 2D RoIs.

\paragraph{RoI pooling features fusion}
\label{fuse} 
We finally fuse the point cloud 3D RoI and the image 2D RoI by aggregating the pooling features along with feature dimension axis as shown in Fig.~\ref{fig:roifusion}. Specifically, we define a fusion strategy by concatenation as in Eq.~\eqref{eq:fusion}:

\begin{equation}\label{eq:fusion}
    \mathbf{F_{fuse}}=MLP(concat[\mathbf{F^{(pc)}_{roi},F^{(img)}_{roi}}])
\end{equation}
where \(\mathbf{F_{fuse}}\) is the fused feature from the 3D RoI pooling features \(\mathbf{F^{(pc)}_{roi}}\) and 2D RoI pooling features \(\mathbf{F^{(img)}_{roi}}\).

\subsection{Prediction layer} The prediction layer, inspired by \cite{yang20203dssd}, use an anchor-free method to directly predict the offset between the center points and corresponding ground truth of the center of the 3D bounding box for regression. Besides, we also directly regress the 3D bounding box size from the fused RoI features. For the orientation regression, we follow the method introduced in \cite{qi2018frustum} that utilizes a hybrid classification and regression formulations to estimate the orientation angle of the 3D bounding box. In particular, we pre-define \(H\) equally split angle bins and use the output of the RoI fusion layer to classify the angle bins, and then regress residual with respect to the classified bin.

\begin{table}[!t]
  \centering
  \caption{3D car detection results on KITTI test dataset. \textit{Sen.} indicates involved sensors by the methods. \textit{L} and \textit{I} denote LIDAR and images respectively.}
  \label{tab:test} \centering
\begin{tabular}{|p{3.5cm}|p{0.5cm}|c|c|c|}
\hline
\multirow{2}{*}{Method} & \multirow{2}{*}{Sen.} & \multicolumn{3}{c|}{\(AP_{Car} (\%)\)}  \\ \cline{3-5} 
                        &                          & easy  & mod. & hard  \\ \hline
\hline
VoxelNet \cite{zhou2018voxelnet}                       & \multirow{9}{*}{L}       & 77.47     & 65.11        & 57.73     \\ \cline{1-1} \cline{3-5} 
SECOND \cite{yan2018second}                       &                          & 83.13     & 73.66        & 66.20    \\ \cline{1-1} \cline{3-5} 
PointPillars \cite{lang2019pointpillars}                      &                         & 79.05    & 74.99        & 68.30   \\ \cline{1-1} \cline{3-5} 
PointRCNN \cite{shi2019pointrcnn}                       &                          & 85.94     & 75.76        & 68.32    \\ \cline{1-1} \cline{3-5} 
Fast PointRCNN \cite{chen2019fast}                      &                         & 84.28     & 75.73        & 67.39    \\ \cline{1-1} \cline{3-5} 
Patches \cite{lehner2019patch}                       &                          & 87.87   & 77.16       & 68.91  \\ \cline{1-1} \cline{3-5} 
Part A2 \cite{shi2020points}                      &                         & 85.94    & 77.86       & 72.00   \\ \cline{1-1} \cline{3-5} 
STD \cite{zhou2018voxelnet}                      &                         & 86.61     & 77.63        & 76.06    \\ \cline{1-1} \cline{3-5} 
3DSSD \cite{yang2019std}                       &                          & 88.36    & 79.57       & 74.55     \\ \hline 
\hline
MV3D \cite{chen2017multi}                       & \multirow{7}{*}{L + I}       & 71.09     & 62.35        & 55.12  \\ \cline{1-1} \cline{3-5} 
AVOD \cite{ku2018joint}                      &                         & 81.94    & 71.88       & 66.38     \\ \cline{1-1} \cline{3-5} 
F-PointNet \cite{qi2018frustum}                       &                          & 81.20    & 70.39       & 62.19  \\ \cline{1-1} \cline{3-5} 
F-ConvNet \cite{wang2019frustum}                      &                          & 85.88     & 76.51        & 68.08   \\ \cline{1-1} \cline{3-5} 
IPOD \cite{yang2018ipod}                      &                          & 79.75     & 72.57        & 66.33  \\ \cline{1-1} \cline{3-5} 
Painted PointRCNN \cite{vora2019pointpainting}                      &                          & 82.11     & 71.70        & 67.08    \\ \cline{1-1} \cline{3-5} 
Ours                      &                          & 88.32     & 79.54        & 74.47    \\ \hline
\end{tabular}
\end{table}

\section{Model structure}

The model structure is presented in Fig.~\ref{fig:structure}. In our experiments we randomly choose \(N=16384\) points from the raw point cloud. We then apply the SA layer Fig.~\ref{fig:sa}, the FP layer Fig.~\ref{fig:fp}, and the image segmentation network \textit{DeepLabv3} to capture \(M=256\) keypoints. The hyper-parameters of the SA layer and the FP layer are represented in Fig.~\ref{fig:sa} and Fig.~\ref{fig:fp} respectively. Successively, we employ a single layer \textit{Votenet} with filters (128) to estimate the center points for the 3D bounding box of the objects. The dimensions of the 3D RoIs are set to \([h=1.8 m,w=5.0 m,l=5.0 m]\), \([h=1.8 m,w=1.0 m,l=1.0 m]\), \([h=1.8 m,w=1.8 m,l=1.8 m]\) for the car, pedestrians, and cyclists objects respectively. We set the constant extended value \(\eta=1.0 m\).

\section{Experiments}

In this section, we evaluate our deep fusion method on the widely used KITTI 3D object detection benchmark \cite{geiger2013vision, geiger2012we}. We firstly introduce the KITTI dataset and explain the detailed training settings. Then, we demonstrate our results by comparison with recent state-of-the-art 3D detectors. We only test our model on the car category due to the large amount of data after preprocessing. However, we evaluate all the categories when we compare our model to the backbone model \textit{3DSSD} \cite{yang2019std} to present the effectiveness of our fusion method. Finally, we analyse the efficiency of our fusion method and visualize some representative results for our 3D object detection model.

\begin{table*}[!t]
  \caption{3D Car detection average precision (AP) on KITTI validation dataset compared to \textit{3DSSD} model. The \textit{Delta} indicates the difference between our model and \textit{3DSSD} model that is our backbone for the point cloud processing. \textit{repro} represents that the results are reproduced on our own computer.}
  \label{tab:delta} \centering
\begin{tabular}{|c|c|c|c|c|c|c|c|c|c|}
\hline
\multirow{2}{*}{Method} & \multicolumn{3}{c|}{\({AP}_{car} \%\)} & \multicolumn{3}{c|}{\({AP}_{ped} \%\)} & \multicolumn{3}{c|}{\({AP}_{cyc} \%\)} \\ \cline{2-10} 
                        & easy   & mod.    & hard  & easy   & mod.    & hard  & easy   & mod.    & hard  \\ \hline
3DSSD (repro)                   & 89.71  & 79.45  & 78.67 & 41.72  & 39.63  & 36.86 & 78.01  & 62.32  & 57.01 \\ \hline
Ours                    & 91.36  & 82.74  & 80.22 & 43.40  & 41.44  & 37.72 & 80.84  & 64.05  & 58.37 \\ \hline
Delta                   & {\color{red} +1.65}   & {\color{red} +3.29}   & {\color{red} +1.55}  & {\color{red} +1.68}   & {\color{red} +1.81}   & {\color{red} +0.86}  & {\color{red} +2.83}   & {\color{red} +1.73}   & {\color{red} +1.36}  \\ \hline
\end{tabular}
\end{table*}

\subsection{Dataset} The KITTI dataset \cite{geiger2013vision} contains both 2D images and 3D point clouds with the corresponding annotations for the cars, pedestrians, and cyclists categories in an urban driving scenario. The sensors used for data collection are: 2 grayscale cameras, 2 color cameras, and 1 Velodyne HDL-64E LIDAR. We only used the point clouds data and images from the left color camera to train our fusion model. The dataset provides 7481 samples for training and 7518 samples for testing. As standard good practice, we further split the KITTI training dataset into 3712 samples for training and 3769 samples for validation. We evaluated our model on the validation dataset following the easy, moderate, and hard difficulty classification levels officially introduced by KITTI. Specifically, in order to align the performance of the algorithms and cover most of the traffic scene scenarios, the object detection task is divided into three levels for validation and testing with respect to the different size, occlusion, and truncation level as shown in Table \ref{tab:levels}. Besides, the average precision (AP) metric is used when we compare our results with other different models.

\subsection{Training settings} We used the Adam \cite{kingma2014adam} algorithm as our training optimizer. The batch size was set to 4 on a NVIDIA 1080Ti GPU. The learning rate was initially set to 0.002, and then was divided by 10 at 40 epochs. Our model has been trained for a total of 50 epochs.

\subsection{Results}

We can firstly compare our model to the backbone network \textit{3DSSD} \cite{yang2019std} on the validation dataset to show the effectiveness of our fusion strategy as shown in Table \ref{tab:delta}. The bottom line indicates the difference between our model and \textit{3DSSD} for 3D car detection. It shows that our model outperforms \textit{3DSSD} in all the categories and all the difficulty levels, which convincingly shows the efficiency of our RoI fusion method.

\begin{figure*}[!t]
  \centering
   \subfigure{\includegraphics[width=0.3\linewidth]{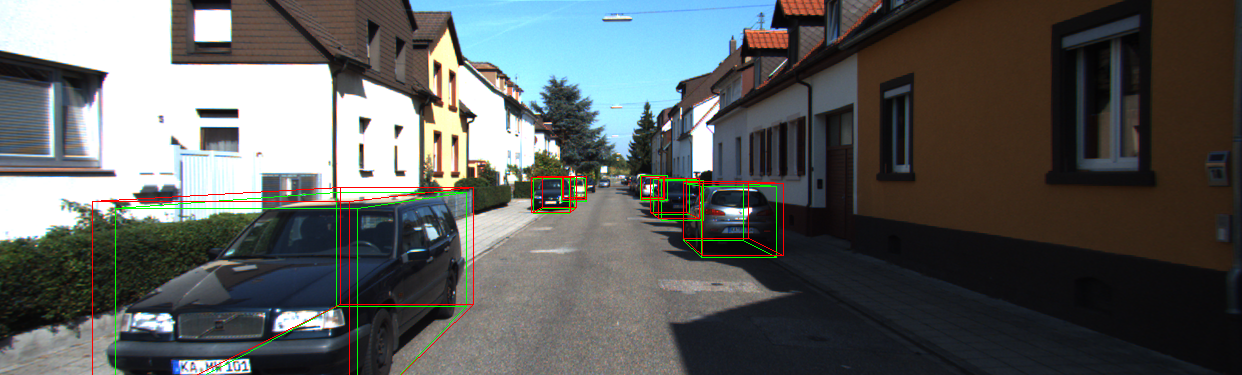}} 
   \subfigure{\includegraphics[width=0.3\linewidth]{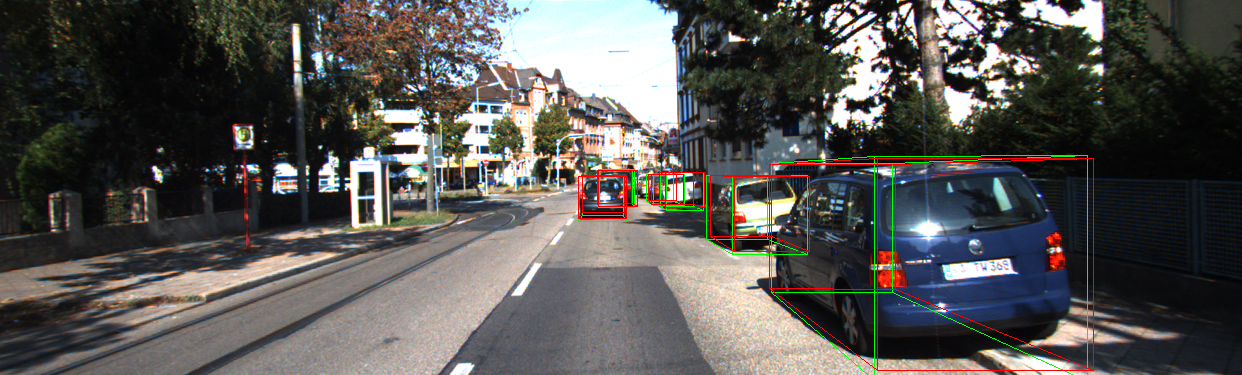}} 
   \subfigure{\includegraphics[width=0.3\linewidth]{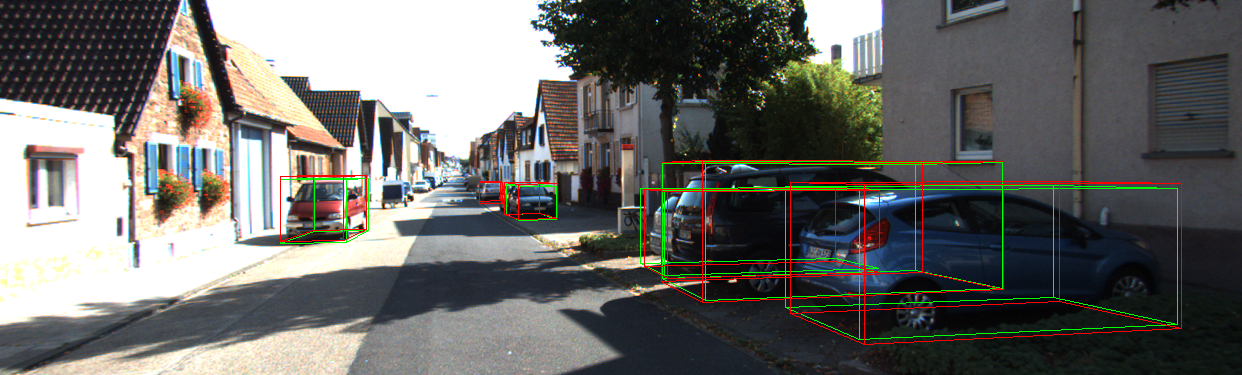}} 
   \subfigure{\includegraphics[width=0.3\linewidth]{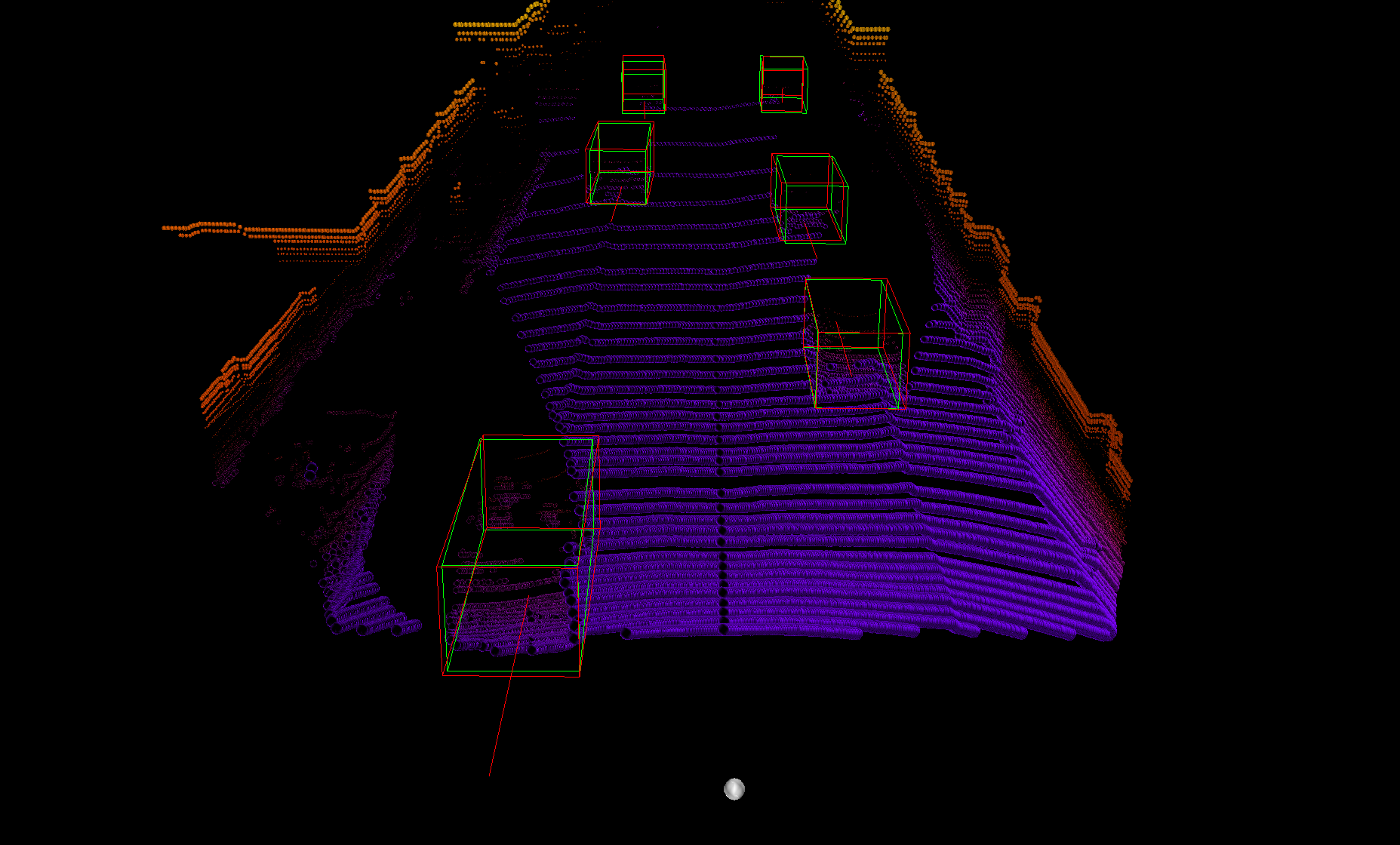}} 
   \subfigure{\includegraphics[width=0.3\linewidth]{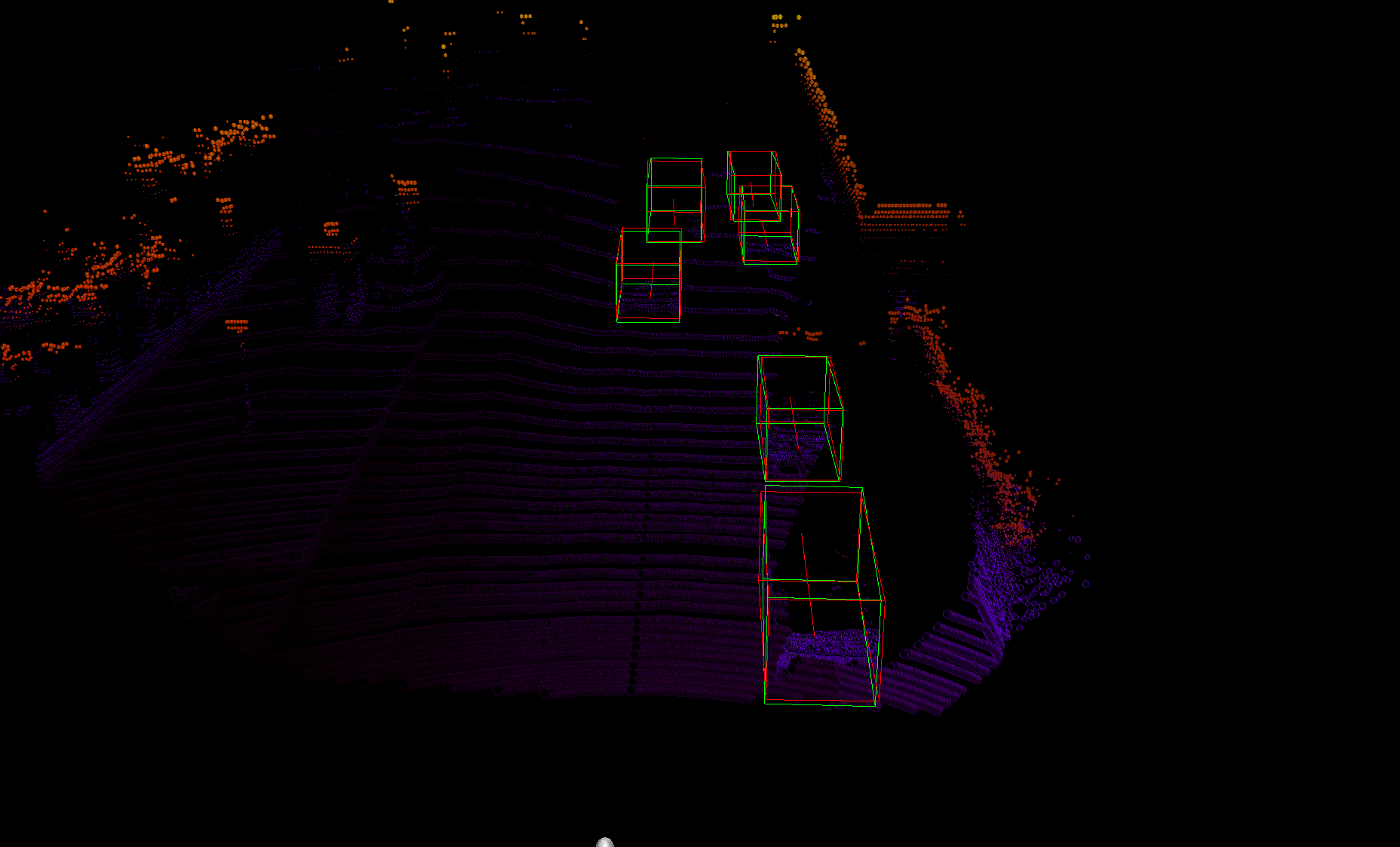}} 
   \subfigure{\includegraphics[width=0.3\linewidth]{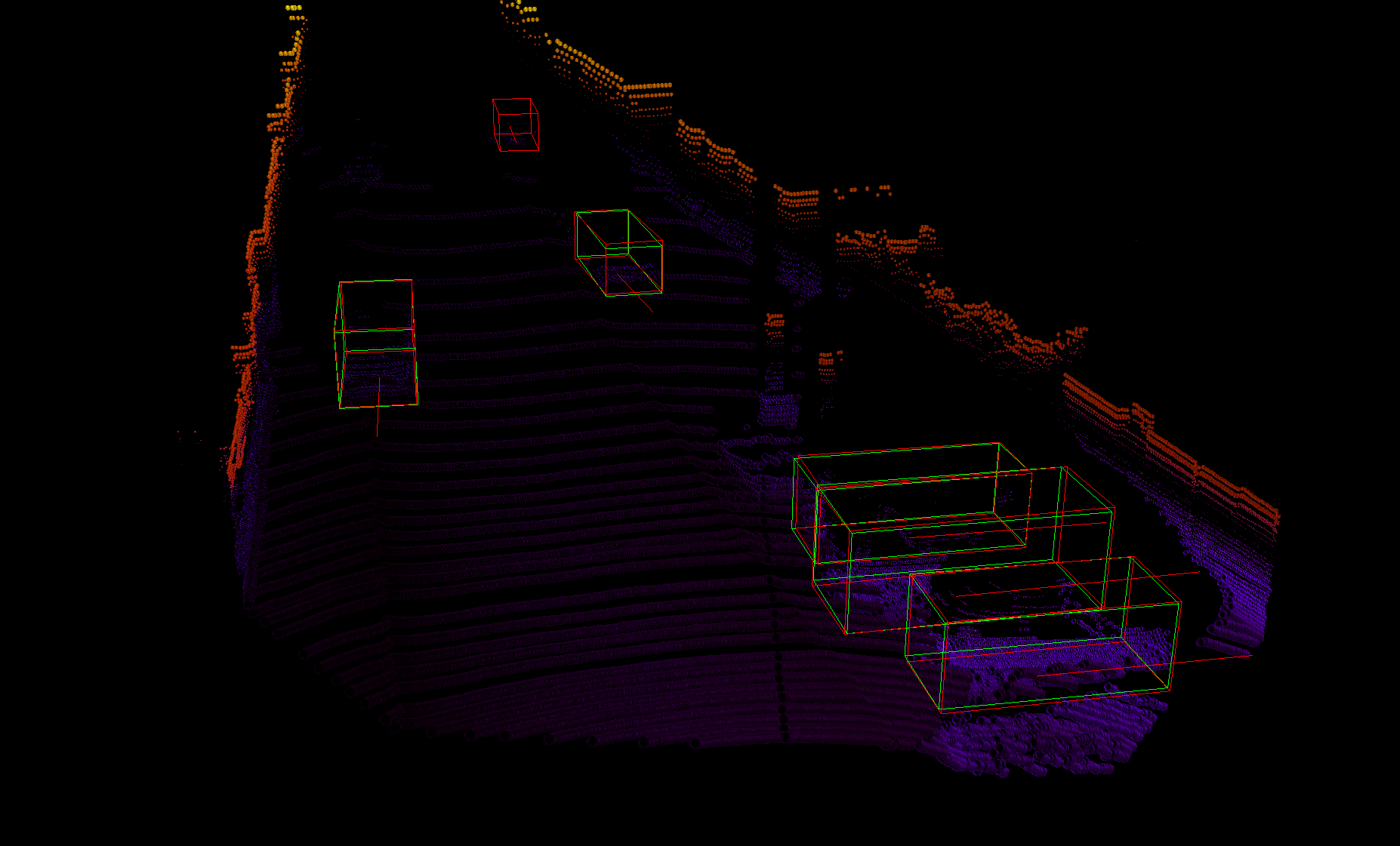}} 
   \subfigure{\includegraphics[width=0.3\linewidth]{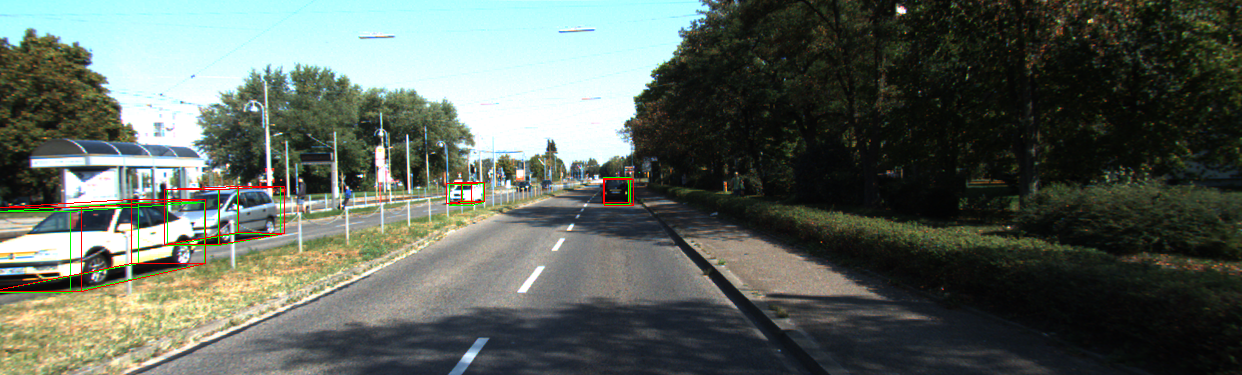}} 
   \subfigure{\includegraphics[width=0.3\linewidth]{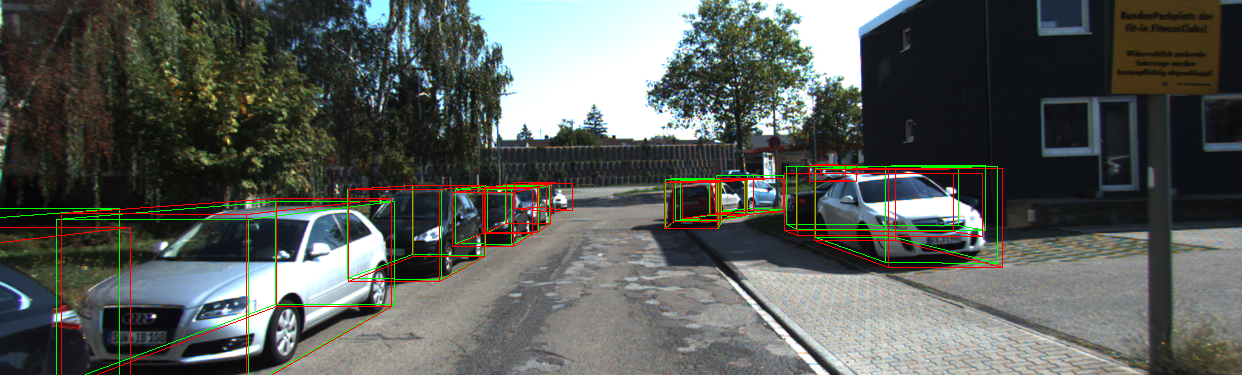}} 
   \subfigure{\includegraphics[width=0.3\linewidth]{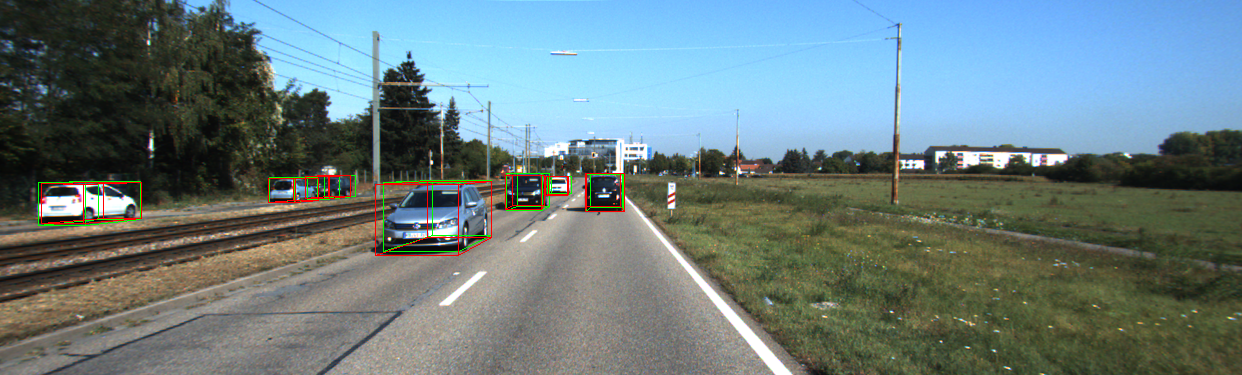}} 
   \subfigure{\includegraphics[width=0.3\linewidth]{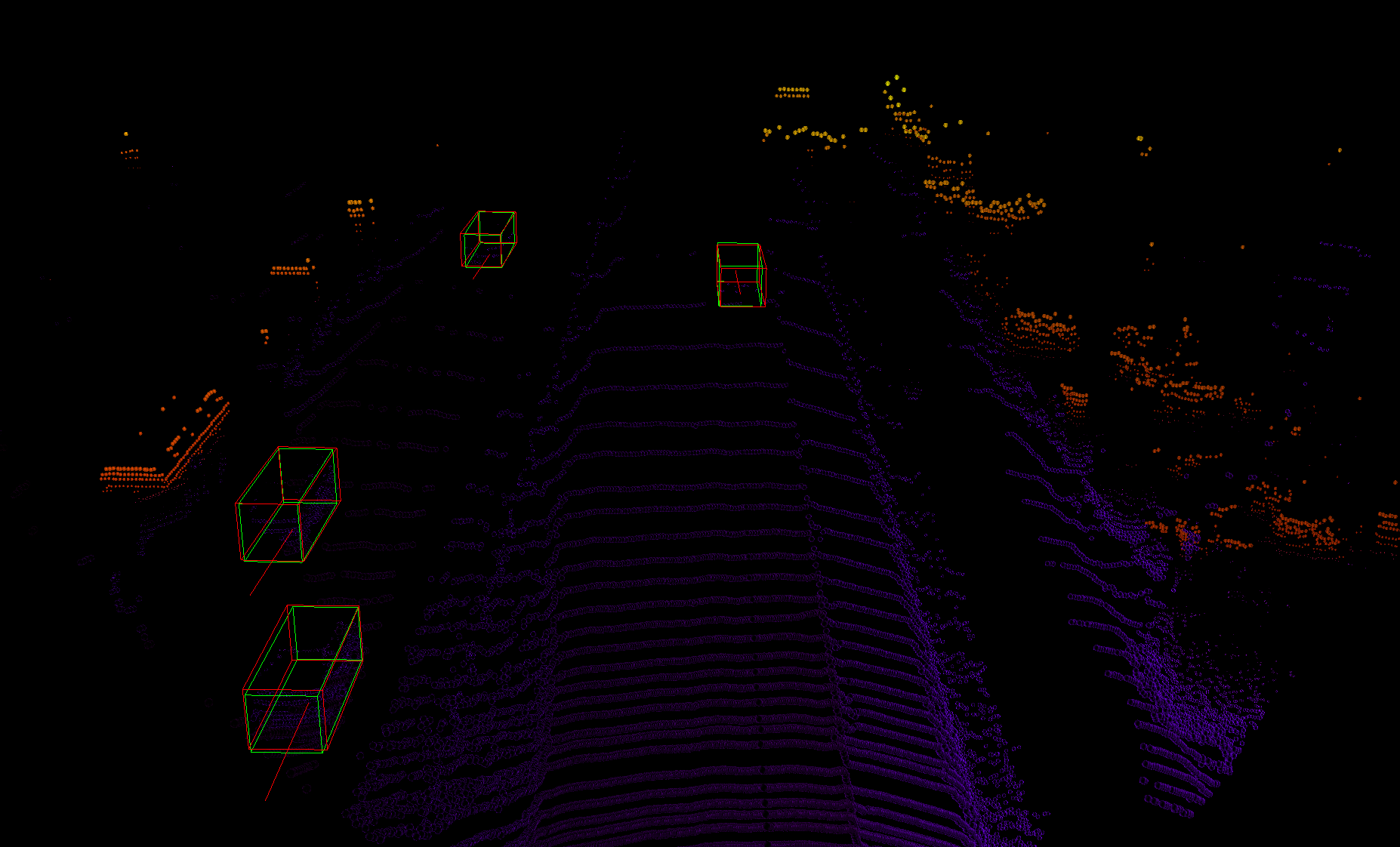}} 
   \subfigure{\includegraphics[width=0.3\linewidth]{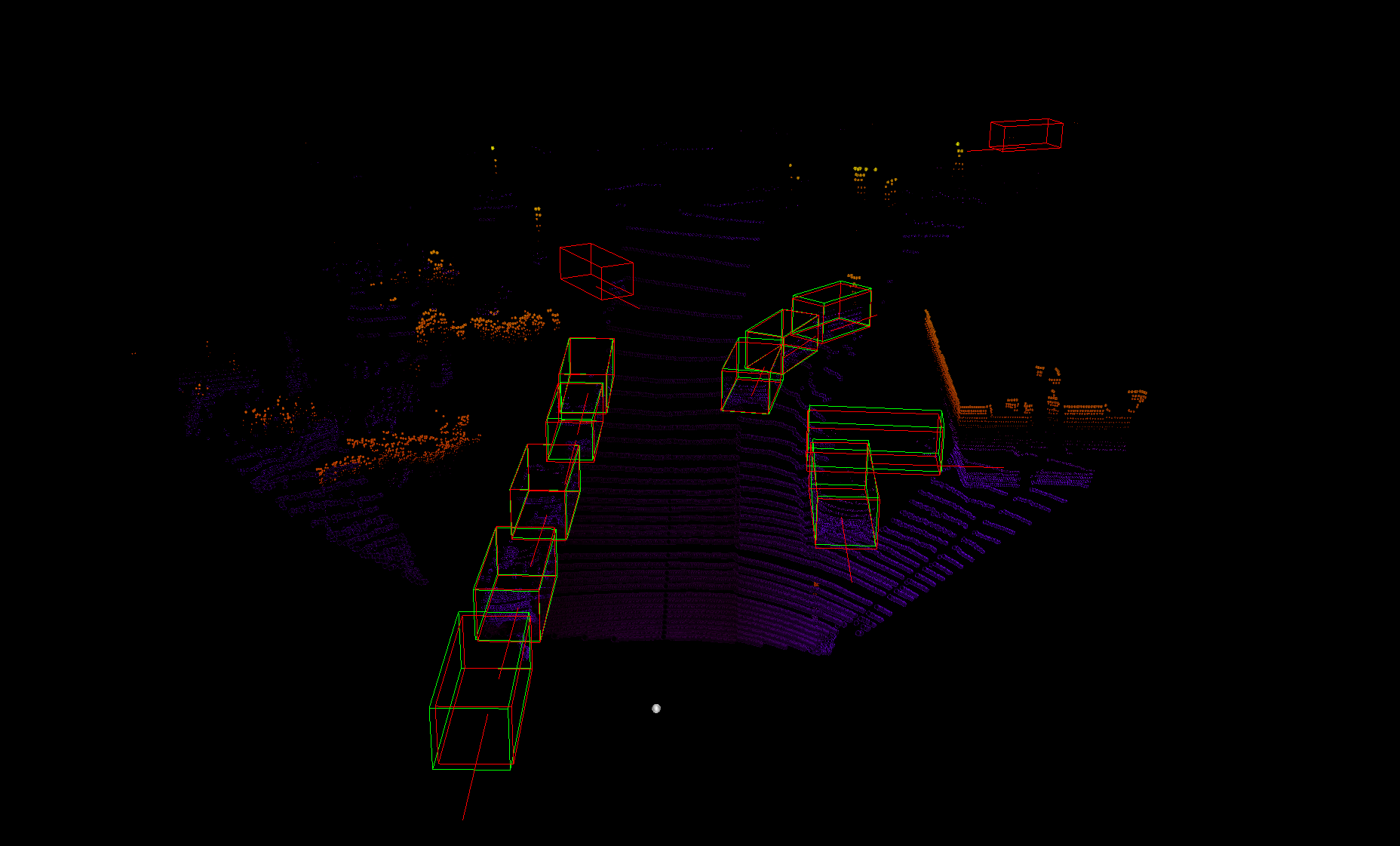}} 
   \subfigure{\includegraphics[width=0.3\linewidth]{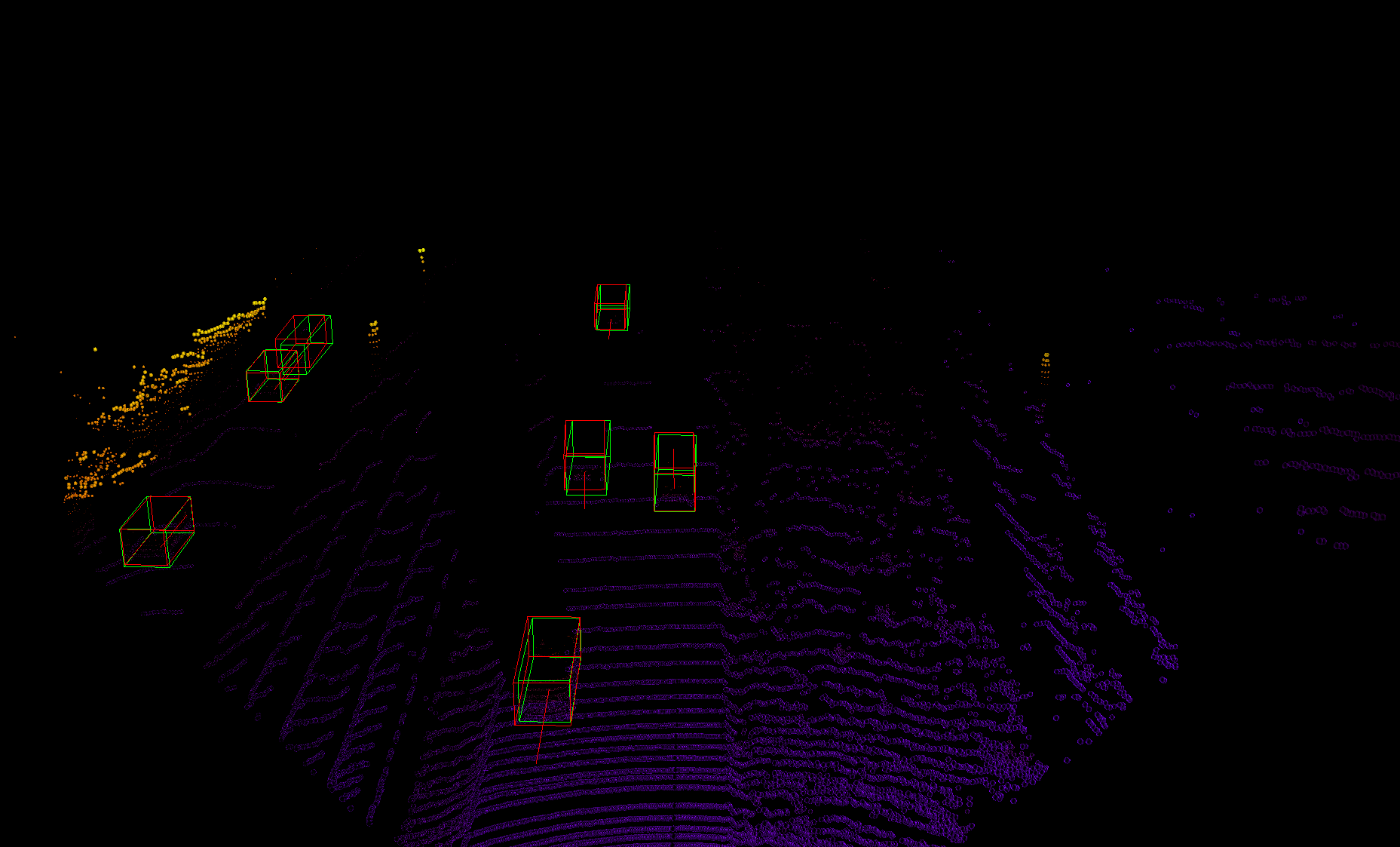}} 
  \caption{Qualitative results on the KITTI validation dataset. The predicted objects and the ground truth objects are shown in red and green bounding boxes respectively. We also project the bounding boxes to the RGB images  for better visualization.}
  \label{fig:visu}
 \end{figure*}

As shown in Table \ref{tab:test}, our model also achieves the best performance compared to recent state-of-the-art fusion methods on the test dataset. We choose \textit{moderate} difficulty as the main average precision (AP) metric, and compare our model to BEV-image fusion \textit{MV3D} \cite{chen2017multi} and \textit{AVOD} \cite{ku2018joint}, our deep fusion method outperforms all others by a large margin. For the frustum method, our method outperforms \textit{F-PointNet} \cite{qi2018frustum} by 9.12\%. Besides, our model significantly outperforms point-pixel-wise fusion method \textit{PointRCNN} \cite{shi2019pointrcnn} by 7.84\%. We also visualize some examples for prediction results and corresponding ground truth as shown in Fig.~\ref{fig:visu} for better representation.

\begin{table}[!]
  \caption{Effectiveness of the extended value \(\eta\) for RoI size on KITTI car validation dataset.}
  \label{tab:size} \centering
  \begin{tabular}{cccc}
    \toprule[0.96pt]
     \(\eta\) (m)  &  \(AP_{easy} (\%)\) & \(AP_{mod.} (\%)\)  & \(AP_{hard} (\%)\) \\
    \midrule
    0.0              & 89.33                & 80.61            & 70.47               \\
    0.5              & 91.12                & 82.53            &79.85               \\
    1.0              & \textbf{91.36}                & \textbf{82.74}  & \textbf{80.22}               \\
    1.5              & 90.42                & 82.48           & 79.11               \\
    2.0              & 89.01                & 81.09            & 78.96               \\
    \bottomrule[0.96pt]
  \end{tabular}
\end{table}

\begin{table}[!]
  \caption{Effectiveness of the fusion strategy on KITTI car validation dataset.}
  \label{tab:fusion} \centering
  \begin{tabular}{cccc}
    \toprule[0.96pt]
     \begin{tabular}[c]{@{}c@{}}fusion \\ strategy \end{tabular}  &  \(AP_{easy} (\%)\) & \(AP_{mod.} (\%)\)  & \(AP_{hard} (\%)\) \\
    \midrule
    sum              & 88.29  & 76.72  & 68.48                \\
    concat              & \textbf{91.36}  & \textbf{82.74}  & \textbf{80.22}               \\
    max              & 87.01  & 74.93  & 67.27                \\
    \bottomrule[0.96pt]
  \end{tabular}
\end{table}

\subsection{Ablation study}

We carried out several ablation experiments to investigate the effectiveness of extended value for RoI size and different fusion methods. All the experiments are performed on the KITTI validation dataset for the 3D car detection task.

\paragraph{Effects of the extended size of RoI} In order to cluster sufficient points around the objects, we enlarge the RoI size by an extended value \(eta\) for more contextual local features. Table \ref{tab:size} shows that the model achieves the best performance when \(\eta=1.0\). Besides, we notice that there is a significant drop of performance when no extended size (i.e. \(\eta=0\)) is used, especially for hard difficulty level of detection. It is assumed that the larger size of the box also provides sufficient information, but is likely to involve more redundant and harmful information. In contrast, smaller size only could provide part information of the cars, which is insufficient to predict the parameters of the 3D bounding box. For the hard detection level objects, they normally are occluded by other objects or far away from the sensor, which leads to very few points on the objects. As a result, involving more surrounding points is beneficial to object classification and regression.

\paragraph{Effects of the fusion strategy} We further investigate the effectiveness for the different fusion strategies. In addition to the concatenation operation as described in Section~\ref{fuse}, we also employ operations, such as sum, max operation, and compare the results for different choices. As shown in Table \ref{tab:fusion}, the concatenation operation for RoI features fusion achieves 91.36\% 82.74\% 80.22\% performance for easy, moderate, and hard difficulties respectively. The results show that the concatenation operation could fuse more discriminative features from the 3D RoIs and corresponding 2D RoIs. This can be linked to the fact that both the sum operation and the max operation could obtain signature features, but the concatenation operation allows to keep all the features from different sensors, which then is likely to allow capturing more useful features for classification and regression.

\begin{figure}[!t]
  \centering
   {\epsfig{file = 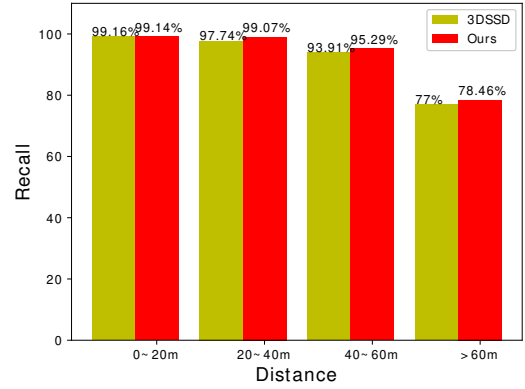, width=0.9\linewidth}}
  \caption{The comparison of the recall for the detected objects in the various distance range.}
  \label{fig:recall}
 \end{figure}
\unskip

\begin{figure}[!t]
  \centering
   {\epsfig{file = 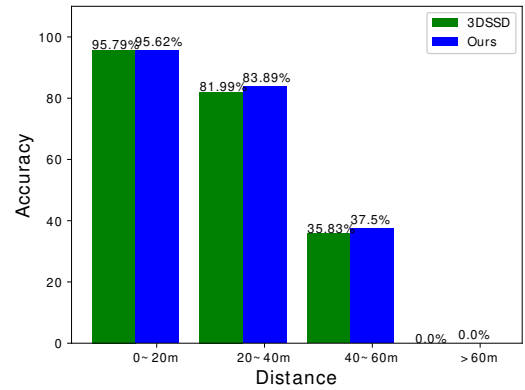, width=0.9\linewidth}}
  \caption{The comparison of the accuracy for the objects in the various distance range.}
  \label{fig:accuracy}
 \end{figure}
\unskip

\paragraph{Effects of the distance of the objects} As we know that LIDAR-only methods are not efficient to detect the objects that include few points, such as small objects and objects in the long distance. In order to better compare our fusion method to the LIDAR-only model \textit{3DSSD} \cite{yang2019std} which only provide car detection model, we detect the cars in the various distance range for the different size of object, and then compare the recall and the accuracy to the \textit{3DSSD}. As illustrated in the Fig.~\ref{fig:recall} and the Fig.~\ref{fig:accuracy}, when the objects are within 20 meters, our results are nearly no different to the \textit{3DSSD}. However, the performance becomes worse when the objects are beyond 20 meters for both models, but our fusion model performs much better than \textit{3DSSD} when detect the objects located in the long distance. The results convincingly prove that our fusion method successfully predicts more true positive objects in the long distance range by learning sufficient colour and texture information for the point clouds.

\begin{table}[!t]\footnotesize
  \caption{Inference time and accuracy on the \textit{moderate} level for different fusion methods.}
  \label{tab:inference} \centering
  \begin{tabular}{|p{0.4cm}|c|c|c|c|}
  \hline
   &  AVOD \cite{ku2018joint} & F-PointNet \cite{qi2018frustum}  & PointPainting \cite{vora2019pointpainting} & Ours  \\ \hline
    time              & 100 ms                & 170 ms            & 400 ms          & 220 ms  \\ \hline
    AP              & 71.48\%                & 70.39\%            & 71.70\%          & 79.54\%  \\ \hline
  \end{tabular}
\end{table}

\paragraph{Inference time} We tested the inference time on KITTI validation dataset with a NVIDIA 1080Ti GPU, and then compare to existing fusion methods in Table~\ref{tab:inference}. Our model achieves the best trade-off compared to BEV-image fusion method \textit{AVOD} \cite{ku2018joint} and frustum method \textit{F-PointNet} \cite{qi2018frustum}. We also note that our RoI fusion method is much better than point-pixel fusion method \textit{PointPainting} \cite{vora2019pointpainting} in terms of both accuracy and inference time.

\subsection{Quantitative analysis}
The comparison of our model with other multi-sensor methods as shown in Table~\ref{tab:test}, shows that our model is more efficient and effective. RoI fusion enables to globally fuse the local area from the point clouds and the images, which make it easier to align the viewpoint when we concatenate the 3D/2D RoI pooling features. In contrast, the point-pixel fusion is unlikely to obtain discriminative features due to the fact that the point clouds are sparse and irregular, but the images distribute as the standard grid. Compared to BEV-image fusion, our model outperforms others by a large margin due to the information losses during BEV generation.

\section{Conclusion}
In this paper, we propose a novel deep fusion method, named RoIFusion, to efficiently fuse the point clouds and the images for 3D object detection. We build a lightweight neural network to generate 3D RoIs from the point clouds and 2D RoIs from the images, and then employ a 3D RoI pooling layer and a 2D RoI pooling layer to obtain the geometric features and the texture features respectively for a potential local area in both point clouds and the images. Finally, we fuse them together to predict the oriented 3D bounding box for the detected object. Our fusion method is flexible and could combine any other LIDAR-only segmentation networks and image segmentation networks. The state-of-the-art performance of our model convincingly show that the fusion method proposed can successfully boost the performance of 3D object detection.

\EOD

\end{document}